\documentclass[letterpaper, 10 pt, conference]{ieeeconf}  

\IEEEoverridecommandlockouts                              
\usepackage{graphicx}

\usepackage{amsmath}
\usepackage{amssymb}
\usepackage{amsfonts}
\usepackage{mathtools}
\usepackage{MnSymbol}
\usepackage{subcaption}
\usepackage{soul}
\usepackage{amsthm}
\usepackage[usenames,dvipsnames,svgnames,table]{xcolor}
\usepackage{float}
\floatstyle{plaintop}
\restylefloat{table}
\usepackage{cite}
\usepackage[normalem]{ulem}

\usepackage{mathtools}
\DeclarePairedDelimiter{\ceil}{\lceil}{\rceil}

\usepackage{mathtools}
\DeclarePairedDelimiter{\floor}{\lfloor}{\rfloor}
\usepackage[vlined,ruled,linesnumbered]{algorithm2e}
\theoremstyle{plain}

\newtheorem{definition}{Definition}

\newtheorem{problem}{Problem}

\newtheorem{example}{Example}
\newtheorem{remark}{Remark}
\newfloat{exlist}{tbp}{lol}
\floatname{exlist}{LIST}

\newcommand{\AP}{{AP}}
\DeclareMathOperator{\luntil}{\mathbf{\mathcal{U}}}

\newcommand{\levent}{\lozenge}
\newcommand{\lalways}{\square}
\newcommand{\rsep}{\mathrel{{.}{.}}\nobreak}
\newcommand{\range}[2]{[{#1 {\rsep} #2}]}

\newcommand{\env}{{Env}}
\newcommand{\States}{{\mathcal{Q}}}
\newcommand{\Transitions}{{\mathcal{E}}}
\newcommand{\Weights}{{\mathcal{W}}}
\newcommand{\Agents}{{\mathcal{J}}}
\newcommand{\Capabilities}{{\mathcal{C}}}

\newcommand{\Resources}{{\mathcal{H}}}

\newcommand{\tavo}[1]{{\color{blue}#1}}

\renewcommand{\baselinestretch}{0.98}

\makeatletter
\let\NAT@parse\undefined
\makeatother
\RequirePackage[bookmarks,unicode,colorlinks=true, allcolors=blue]{hyperref}%

\title{\LARGE \bf
An Iterative Approach for Heterogeneous Multi-Agent Route Planning with Resource Transportation Uncertainty and Temporal Logic Goals}

\author{Gustavo A. Cardona$^*$, Kaier Liang$^*$, and Cristian-Ioan Vasile
\thanks{$^*$ The authors contributed equally.}
\thanks{Gustavo A. Cardona, Kaier Liang, and Cristian-Ioan Vasile  are with the Mechanical Engineering and Mechanics Department at Lehigh University, PA, USA: {\tt\small \{gcardona, kal221, cvasile\}@lehigh.edu}}        
}     

\begin{document}
\maketitle
\thispagestyle{empty}
\pagestyle{empty}

\begin{abstract}
This paper presents an iterative approach for heterogeneous multi-agent route planning in environments with unknown resource distributions. 
We focus on a team of robots with diverse capabilities tasked with executing missions specified using Capability Temporal Logic (CaTL), a formal framework built on Signal Temporal Logic to handle spatial, temporal, capability, and resource constraints. 
The key challenge arises from the uncertainty in the initial distribution and quantity of resources in the environment. 
To address this, we introduce an iterative algorithm that dynamically balances exploration and task fulfillment.
Robots are guided to explore the environment, identifying resource locations and quantities while progressively refining their understanding of the resource landscape. 
At the same time, they aim to maximally satisfy the mission objectives based on the current information, adapting their strategies as new data is uncovered. 
This approach provides a robust solution for planning in dynamic, resource-constrained environments, enabling efficient coordination of heterogeneous teams even under conditions of uncertainty.
Our method's effectiveness and performance are demonstrated through simulated case studies. 
\end{abstract}

\section{Introduction}
\label{sec:introduction}

Using diverse multi-robot systems with unique capabilities is becoming increasingly crucial for complex real-world tasks like disaster response, autonomous construction, and planetary exploration~\cite{cardona2019robot, gregory2016application, cardona2022planning, weisbin2000nasa}. 
In such scenarios, combining the specialized skills of different robots can greatly enhance task efficiency, resilience, and robustness. 
By working together, robots with expertise in areas like aerial surveillance, manipulation, or drilling can achieve mission goals that would be unattainable for uniform systems~\cite{di2010autonomous,parascho2023construction,zaheer2016aerial,fang2024continuous}. 
This varied range of capabilities enables flexible task assignments, allowing robots to capitalize on their strengths and overcome challenges such as environmental uncertainty and time limitations~\cite{jing2018accomplishing,jing2016end, buyukkocak2023energy}. 
As missions grow in complexity, establishing formal frameworks that account for the diverse nature of robot capabilities and task coordination becomes essential to ensure system effectiveness and scalability.

In recent years, temporal logic has emerged as a powerful formalism for specifying and planning tasks in multi-robot systems, enabling the expression of complex spatial, temporal, and logical constraints. 
Linear Temporal Logic (LTL) has been widely used for coordinating robots by specifying task order, but its lack of explicit time constraints makes it less suitable for time-sensitive missions~\cite{tumova2016multi,kantaros2020stylus,kress2009temporal}. 
Signal Temporal Logic (STL) addresses this limitation by allowing reasoning over both explicit time and task execution~\cite{sun2022multi,gundana2021event,lindemann2019coupled,sewlia2023maps, buyukkocak2021planning}. 
Capability Temporal Logic (CaTL)~\cite{Jones2019ScRATCHS, leahy2021scalable,leahy2022fast}, a syntactic extension of STL, was developed to handle heterogeneous robots by incorporating their diverse capabilities into task specifications. 
CaTL was later extended in \cite{cardona2024planning} to model resource transportation, enabling the planning of both agent actions and the movement of divisible and indivisible resources. 
However, none of these previous approaches can effectively address uncertainties arising from the environment, task execution, or robot actions, which can compromise mission feasibility.
\begin{figure}[t]
    \centering
    \includegraphics[width=\linewidth]{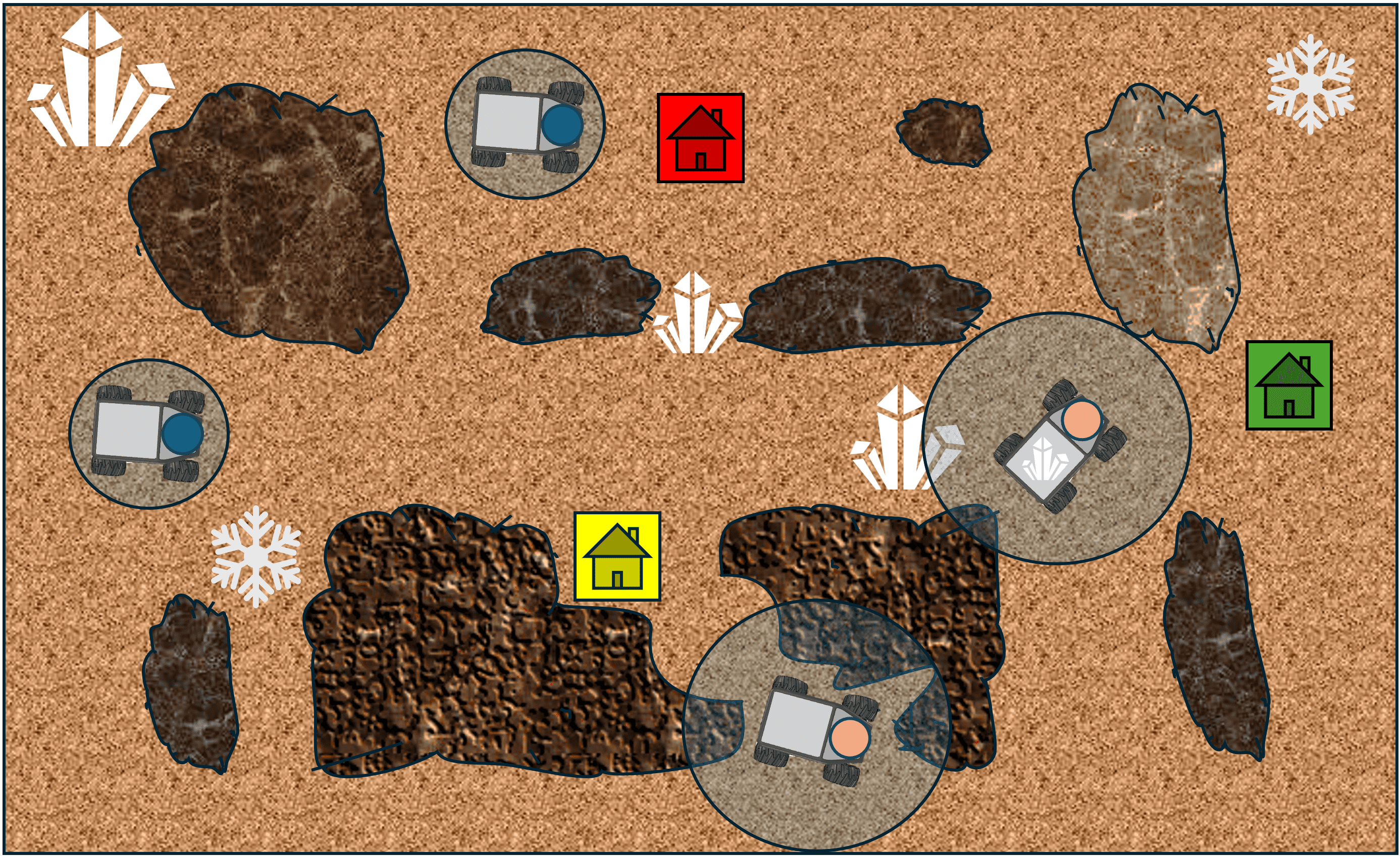}
    \caption{Example of heterogeneous robot planning on a planetary exploration mission. Two robot classes are looking for ice and minerals.}
    \label{fig:problem_example}
\end{figure}

Uncertainty in such types of problems with multi-robot systems has been addressed by introducing probabilistic temporal logics~\cite{sadigh2016safe, schillinger2018auctioning,guo2023hierarchical}, which incorporate probabilities into task specifications to account for unpredictable events. 
Additionally, chance constraints have been used to handle uncertainties in perception and robot travel time, allowing robots to make decisions that maximize the likelihood of task success within acceptable risk levels~\cite{jha2018safe, cai2021probabilistic, liang2024iterative}. 
This paper addresses the problem of planning for CaTL specifications in environments where the distribution and quantity of resources are initially unknown. 
We propose an iterative algorithm that balances exploration for resource discovery to maximally satisfy the mission based on the available information. 
Furthermore, we account for uncertainty in resource detection due to sensor limitations and implement a progressive refinement of the resource distribution belief, enhancing the accuracy of the system's understanding over time and ensuring robust task planning even when facing incomplete information or variable conditions.

The main contributions of this work are, 
\begin{enumerate}
    \item We propose a heterogeneous multi-robot route planning problem for tasks requesting specific capabilities and resources while considering unknown resource distribution in the environment. We impose robots to discover resources and estimate their amount depending on the agent's capability.
    \item We propose an iterative algorithm approach that combines partial mission satisfaction with current knowledge of resource distribution and the progressive refinement of resource distribution belief.
    \item We extend the MILP-encoding of CaTL~\cite{cardona2024planning},  to encourage agents to explore the environment while simultaneously satisfying the mission from up to date information about resource distribution.
    \item We evaluate the time performance of our planning approach in practical scenarios, such as planetary exploration. By varying the environment size and the number of agents, we aim to gain insights into the problem's characteristics and its application in real-world situations.

\end{enumerate}

\noindent
\textbf{Notation:} Let $\mathbb{Z}$, $\mathbb{R}$, and $\mathbb{B}$ denote the sets of integers, real numbers, and binary values, respectively. The sets of integers and real numbers greater than or equal to $a$ are denoted by $\mathbb{Z}_{\geq a}$ and $\mathbb{R}_{\geq a}$, respectively. For any set $\mathcal{S}$, $2^\mathcal{S}$ represents its power set, and $|\mathcal{S}|$ denotes its cardinality. The notation $x \sim \mathcal{X}$ indicates that the random variable $x$ is drawn from the probability distribution $\mathcal{X}$. For a set $\mathcal{S} \subseteq \mathbb{R}$ and a scalar $\alpha \in \mathbb{R}$, $\alpha + \mathcal{S}$ denotes the set $\{\alpha + x \mid x \in \mathcal{S}\}$. The empty set is represented by $\emptyset$. Let $\range{a}{b} = \mathbb{Z} \cap [a, b]$ represent the set of integers between $a$ and $b$ (inclusive). For a range $I = \range{a}{b}$, we use $\underline{I} = a$ and $\bar{I} = b$ to refer to the lower and upper bounds, respectively.



\section{Problem Formulation}

Here, we introduce a route planning problem for heterogeneous multi-robot teams operating in environments with an unknown distribution of resources. 
The robots are tasked with fulfilling missions specified using Capability Temporal Logic (CaTL)~\cite{Jones2019ScRATCHS, leahy2021scalable}, which incorporates tasks requesting specific agent capabilities and then extended in \cite{cardona2024planning} to accommodate resource constraints. 
Depending on their class, the robots are equipped with sensors of varying accuracy and range to explore and identify resource locations. 
The accuracy of each sensor determines the team's confidence in estimating the quantity of resources within a sensed area, influencing decision-making during the mission. 
We extend the CaTL framework in \cite{cardona2024planning} to accommodate this dynamic resource discovery process, allowing the team to refine its understanding of resource availability while optimizing the route planning for resource transportation and capabilities of agents.
Moreover, the robot is required to perform the mission repeatedly, e.g., daily.
We formalize the problem as an optimal route planning problem, where robots must efficiently explore the environment to gain more information about resources and maximize task satisfaction over the mission iterations.

The following example illustrates the problem class addressed in this paper, where resources are needed at specific locations in the environment for tasks to be completed by agents with diverse capabilities.

\begin{example}
In the scenario shown in Fig.~\ref{fig:problem_example}, multiple robots are equipped with various capabilities, such as cameras, soil sample extractors, drilling tools, and spectrometers. 
These robots are tasked with exploring a planet, searching for ice and minerals to collect and store in specific headquarters, all while avoiding craters.
Each robot belongs to a specific class based on its combination of capabilities (e.g., a robot with camera and drilling capabilities). 
These robot classes can have cameras with different sensing ranges and resource detection accuracies, which can affect the precision and reliability of their measurements. 
Furthermore, each robot class may have a different storage capacity for transporting resources.
A planetary exploration task aims to locate a certain amount of resources while ensuring that the necessary capability is available at a specific location. 
However, due to the uncertainty of discovering the resources and the distribution of agents, tasks may not be fulfilled simultaneously, in sequence, or in full.

As an illustration, a list of tasks for a planetary exploration mission may look like this:
\begin{exlist}[ht]
\hrulefill
\vspace{-2mm}
\begin{enumerate}
    \item From deployment to the end of the day, one video surveillance is required at the red headquarters. 
    \item From deployment to the end of the day, 200 grams of ice and a drilling capability are required at Yellow headquarters. 
    \item Within 4 and 12 hours after deployment, a spectrometer and 10 grams of mineral are required in every headquarters for 1 hour.
    \item Within 12 hours after deployment to the end of the day, 2 sample extractors and 200 grams of either ice or minerals are required for 6 hours at Green headquarters.
\end{enumerate}
\vspace{-4mm}
\hrulefill
\caption{Example list of planetary exploration tasks requiring capabilities and resources.}
    \label{listOfTasks}
\end{exlist}
%
\end{example}
Based on this example, we define the environment, capabilities, resources, and robot models.

\subsection{Environment, robot, and resource models}
We consider a group of heterogeneous robots, denoted by $\Agents$, operating in an environment represented by a transition system $\env = (\States, \Transitions, \Weights, \AP, \mathcal{L})$. 
The set $\States$ consists of states $q$, which correspond to locations of interest in the environment, and $\Transitions \subseteq \States \times \States$ defines the possible transitions between these locations. 
The function $\Weights: \Transitions \rightarrow \mathbb{Z}_{\geq 1}$ assigns a positive integer to each transition, representing its travel time, based on a discretization of time $\Delta k > 0$. 
A stationary robot at a location is described by the transition $(q, q) \in \Transitions$, where \(\mathcal{W}(q, q) = 1\). 
Each state is associated with an atomic proposition $\pi \in \AP$, determined by the labeling function $\mathcal{L}: \States \rightarrow \AP$. 
The set of all states labeled with a specific atomic proposition $\pi$ is given by $\mathcal{L}^{-1}(\pi) = \{q \mid \mathcal{L}(q) = \pi\}$.

$\Resources$ denotes the set of all resources distributed in the environment, and resources are considered, \emph{divisible} (e.g., such as sand and water) and \emph{indivisible} (e.g., such as bricks and wooden beams). 

\begin{remark}
    For simplicity, we assume that the  amount of resources and distribution in the environment
     resets at the start of each task execution.
    When agents complete a task, the resources they have consumed or collected during that task are replenished to their original levels.
    However, our approach can handle (possibly unknown) resource dynamics associated with consumption and production.
    
\end{remark}

Each robot $j \in \Agents$ has a set of capabilities $c_j$ which defines its robot class, $g = c_j$, drawn from a finite set of capabilities $\Capabilities$. The set of all robot classes is denoted by $\mathcal{G} \subseteq 2^{\Capabilities}$. 
We consider every robot to have at least a camera sensor $c_0$, represented as capability $ c_0 = \boldsymbol{\sigma}\in \mathbb{R}^{|\Resources|}$, where each component represents the standard deviation for detecting a specific type of resource. 
For a robot of class $g \in \mathcal{G}$ and a resource $h \in \Resources$, 
the measurement accuracy is characterized by the standard deviation $\sigma_{g,h}$, which corresponds to the $h$-th component of $\boldsymbol{\sigma}$ for the camera sensor in class $g$.

We also consider robots capable of transporting resources as defined in \cite{cardona2024planning}.  
Each robot class $g \in \mathcal{G}$ is assigned a storage capacity that dictates the amount of resources it can transport. 
Similarly, we consider the distribution of resource $h\in\Resources$ in the environment denoted by $\mathcal{R} \in \mathbb{R}^{|\Resources| \times |\mathcal{Q}|}$ where $|\Resources|$ is the number of resource types and $|\States|$ is the number of states (locations) in the environment.

We consider the initial distribution of resources to be unknown. To represent our belief about resource distribution, we introduce a belief resources map $\tilde{\mathcal{R}} \in \mathbb{R}^{|\Resources| \times |\States|}$,  Initially, $\tilde{\mathcal{R}}$ is set to zero ($\tilde{\mathcal{R}} = \mathbf{0}$), representing no prior information about the environment's resource distribution.
The estimated distribution of resources discovered by robots at any time $t$ is captured by the belief resources map $\tilde{\mathcal{R}}_t$, where $\tilde{\mathcal{R}}_t(q) \in \mathbb{R}^{\Resources}{\geq \boldsymbol{0}}$ represents a vector of different resource $h$ believed to be available at state $q$ at time $t$. 
We consider agents pick-up a deterministic amount of resources from the environment according to the belief of that state, except when the true value is less than the belief.


Resource exchanges are permitted only at designated states $q \in \States$, ensuring adherence to mission requirements.  

A trajectory on the environment for agent $j \in \Agents$ is denoted by the mapping function $s_j : \mathbb{Z}_{\geq 0} \to \States \cup \Transitions$. 
Here, $s_j(k)$ indicates whether robot $j$ is located at state $q \in \States$ or moving along edge $e \in \Transitions$ at time $k \in \mathbb{Z}_{\geq 0}$, with the initial condition $s_j(0) = q_{0, j}$, where $q_{0, j} \in \States$ is the initial location of robot $j$.
The synchronous trajectory of a set of agents $\Agents$ is 
$s_\Agents: \mathbb{Z}_{\geq0} \to (\States \cup \Transitions)^{|\Agents|}$.
The number of robots at state $q \in \States$ with capability $c \in \Capabilities$ at time $k \in \mathbb{Z}_{\geq 0}$ is captured by the variable $n_{q, c}(k) = |\{j \in \Agents \mid q=s_j(k), c \in c_j\}|$.
Similarly as for agent we consider the distribution of resource $h\in\Resources$ in the environment denoted by $b_h: \mathbb{Z}_{\geq0} \times (\States \cup \Transitions) \to \mathbb{R}_{\geq 0}$.
The total distribution of resources picked up by robots and therefore be known to exist are denoted by $b_\Resources$.

\subsection{Capability Temporal Logic (CaTL)}
Here, we define the syntax and semantics of Capability Temporal Logic (CaTL), a formal language designed for coordinating a heterogeneous team of robots tasked with transporting resources. 
Within CaTL, tasks are the fundamental units of a specification, as detailed below.
\begin{definition}[Task~\cite{cardona2024planning}]
\label{def:task}
A \emph{task} is a tuple $T = (d,\pi, cp, rs)$,
where $d \in \mathbb{Z}_{\geq 1}$ is a discrete duration of time (i.e., multiple time steps $\Delta k$), $\pi\in \AP$ is an atomic proposition, $cp: \Capabilities \to \mathbb{Z}_{\geq 0}$ and $rs: \Resources \to \mathbb{R}_{\geq 0}$ are counting maps specifying how many agents with each capability and how many resources of each type should be in each region labeled $\pi$, respectively.
Note that for divisible resources $rs(h) \in \mathbb{R}_{\geq 0}$, while for indivisible ones $rs(h) \in \mathbb{Z}_{\geq 0}$. 
\end{definition}

\begin{definition}
\label{def:catl-syntax}
The \emph{syntax} of CaTL~\cite{leahy2021scalable} is a fragment of Signal Temporal Logic (STL)~\cite{maler2004}
\begin{equation*}
\label{eq:catl_syntax}
    \phi ::=  T \mid \phi_1 \land \phi_2 \mid \phi_1 \lor \phi_2 \mid \phi_1 \luntil_{I} \phi_2 \mid \levent_{I} \phi \mid \lalways_{I} \phi
\end{equation*}
where $\phi$ is a CaTL formula, $T$ is a task, $\land$ and $\lor$ are the Boolean conjunction and disjunction operators, $\luntil_{I}$, $\levent_{I}$, and $\lalways_{I}$ are the time-bounded until, eventually, and always operators, respectively, with $I=[a \rsep b]$ a time interval.
\end{definition}

Similar to STL, CaTL also has qualitative semantics recursively defined as follows.

\begin{definition}[CaTL qualitative semantics~\cite{cardona2024planning}]
\label{def:catl-semantics}
The Boolean (qualitative) semantics of CaTL are defined over trajectories $s_\Agents$ of agents and $b_\Resources$ of resources.
At time $k$, the semantic is 
\begin{equation}
{\small
    \begin{aligned}
    (s_\Agents, b_{\mathcal{H}},k) &\models T \Leftrightarrow  \forall \tau \in [k \rsep k+d], \forall q \in L^{-1}(\pi),\\
    &\qquad \quad\forall c \in cp_T, \forall h \in rs_T,\\
    &\qquad \quad n_{q, c}(\tau) \geq cp(c)\, \land\, b_h(k, q) \geq rs(h) \:.\\
    (s_\Agents, b_{\mathcal{H}}, k) &\models \phi_1 \lor \phi_2 \equiv{}   \big( (s_\Agents, b_{\mathcal{H}}, k)\models \phi_1 \big) \lor \big( (s_\Agents, b_\mathcal{H}, k)\models \phi_2 \big),\\
    (s_\Agents, b_{\mathcal{H}}, k) &\models \phi_1 \land \phi_2 \equiv{}   \big( (s_\Agents, b_{\mathcal{H}}, k)\models \phi_1 \big) \land \big( (s_\Agents, b_\mathcal{H}, k)\models \phi_2 \big),\\
    (s_\Agents, b_{\mathcal{H}}, k) &\models \levent_{I} \phi \equiv{} \exists k' \in k+I  \textrm{ s.t. } (s_\Agents, b_\mathcal{H},k') \models \phi,\\
    (s_\Agents, b_{\mathcal{H}}, k) &\models \lalways_{I} \phi \equiv{} \forall k' \in k+I  \textrm{ s.t. } (s_\Agents, b_\mathcal{H},k') \models \phi.
\end{aligned}
}
\end{equation}
A pair of team and resource trajectories satisfy a CaTL formula $\phi$,
denoted $(s_\Agents, b_\Resources, k) \models \phi$ if $(s_\Agents, b_\Resources, 0) \models \phi$.
\end{definition}

\subsection{Problem}
We formally define the problem of heterogeneous robot route planning for discovering resources and satisfy a CaTL specification repeatedly as follows
\begin{problem}
\label{pb: pb_1}
Given a set of robots $\Agents$ with initial states $q_{0,j}$, capabilities $c_j \subseteq \Capabilities$ deployed in environment $Env$ with an unknown initial distribution of resources $\mathcal{H}$, and a CaTL specification $\phi$ that must be satisfied repeatedly, find trajectories $s_j$ for agents to explore and transport resources $h \in \Resources$ such that $(s_\Agents, b_\Resources)$ satisfy the task $\phi$ as much as possible in each execution iteration.
\end{problem}

\section{Solution}
\label{sec:Solution}
This section proposes an iterative approach for multi-robot task planning in environments with uncertain resource distributions in Pb.~\ref{pb: pb_1}. Our method combines task execution with progressive refinement of resource beliefs. The structure of the iterative algorithm is shown in Alg.~\ref{alg: iterative}. It consists of two main phases: a planning phase (line 6) that computes the optimal assignment and planning based on the current belief and an updating phase (line 7) that refines the belief based on the computed assignment.

The algorithm maintains and updates two key components: The first is belief map $\tilde{\mathcal{R}} \in \mathbb{R}^{|\Resources| \times |\mathcal{Q}|}$, where $|\Resources|$ is the number of resource types and $|\mathcal{Q}|$ is the size of the map (i.e., number of locations). Initially, $\tilde{\mathcal{R}}$ is set to zero ($\tilde{\mathcal{R}} = \mathbf{0}$), representing no prior information about the environment's resource distribution.

The second is uncertainty map $\Omega \in \mathbb{R}^{|\mathcal{Q}|}$, which quantifies the relative confidence level for each location in the environment. For each location $q \in \mathcal{Q}$, $\Omega(q) \in (0, 1)$, with lower values indicating higher confidence. The uncertainty degrees are normalized such that $\sum_{q \in \mathcal{Q}} \Omega(q) = 1$. 
Initially, we set $\Omega(q) = \frac{1}{|\mathcal{Q}|}$ for all $q$, representing uniform uncertainty across the environment. As the algorithm progresses, $\Omega$ will be updated based on the team's assignment and observations, with its values derived from normalized variances of our belief about resource distribution explained later.

As the iterations progress, both $\tilde{\mathcal{R}}$ and $\Omega$ are updated based on the agents' observations, gradually refining our understanding of the resource distribution and reducing uncertainty in explored areas. 


\begin{algorithm}
\caption{Iterative approach}
\label{alg: iterative}
$\Agents$ -- Agents, $\phi$ -- task specification,\\
$\mathcal{R}$ -- ground truth, $\Tilde{\mathcal{R}}$ --  initial belief map\\
$\Omega $ --  uncertainty map
\DontPrintSemicolon
\BlankLine
\small
$\Tilde{\mathcal{R}} \gets \boldsymbol{0}$, $\Omega \gets \boldsymbol{\frac{1}{|\States|}}$\\
\While{\texttt{Check\_Criteria}}{
$plan \gets \texttt{ComputePlan}(\Tilde{\mathcal{R}}, \phi, \Agents, \Omega$)\\
$obs \gets \texttt{Execute}(plan)$\\
$\Tilde{\mathcal{R}}, \Omega = $ \texttt{UpdateBelief($\Tilde{\mathcal{R}}, \Omega, obs$)}\\
}
\end{algorithm}

The iterative process continues until a termination condition (\texttt{Check\_Criteria}) is met in line 5. This condition can be triggered either when a pre-defined maximum number of iterations is reached or when the task satisfaction fraction executed based on the current $plan$ converges to a stable value. 
Each iteration comprises three steps. 
In the plan generation step (\texttt{ComputePlan}, line 6), the algorithm creates a $plan$ for each agent, optimizing the current belief to best satisfy specification $\phi$. 
During plan execution (\texttt{Execute}, line 7), agents carry out the $plan$, collecting a sequence of observations $obs$ from sensors in the real system. 
The belief update step (\texttt{UpdateBelief}, line 8) then refines the belief and uncertainty maps based on these new observations. 

We do not consider mission failure or online adaptation due to the overestimation of required resources during the task execution.
We plan to deal with this in future work in a similar way as we tackled the probability of success in \cite{liang2024iterative}.



\subsection{Task satisfaction and environment exploration}
\subsubsection{Satisfaction}
We encode the CaTL specification to satisfy tasks by translating it into an STL specification.
This is possible since CaTL is a fragment of STL (CaTL $\subseteq$ STL).
CaTL allows intuitive, compact encodings of tasks, and like STL, it can be efficiently encoded as a MILP~\cite{leahy2021scalable,cai2021probabilistic, liu2023robust,cardona2022partial}.
However, in this work we translate the CaTL specification into STL by transforming tasks.
A task $T = (d,\pi, cp, rs)$ defined as in Def. \ref{def:task} is semantically equivalent to the following STL specification formula
$$\phi_T = \bigwedge_{h \in \Resources} \varsigma(\pi, h) \land \lalways_{[0,d]} \bigwedge_{c \in \Capabilities} \varsigma(\pi, c),$$
where $\varsigma(\pi, c)$ and $\varsigma(\pi, h)$ are 
\begin{align}
    \label{eq:thresshold capacities}
    \varsigma(\pi, c) &= \min_{q \in L^{-1} (\pi)} \{n_{q,c}\} \geq cp(c),\\
    \label{eq:thresshold resources}
    \varsigma(\pi, h) &= \min_{q \in L^{-1} (\pi)} \{b_h(\cdot, q)\} \geq rs(h).
\end{align}
Then, as the resource distribution can be partially or complete unknown, it needs to partially satisfy the mission specification. 
For this we use our encodings that captures partial satisfaction for STL specifications via recursive encoding using a variable $z_k^{\phi} \in [0,1]$ for Boolean and temporal operators and $z_k^\mu \in \mathbb{B}$ for predicates.
The complete encoding follows similar to the one in \cite{cardona2022partial, cardona2023preferences}, and we omit it for brevity.
The partial satisfaction allows the multi-agent system to execute tasks resulting in a satisfaction fraction $J_{\text{satisfaction}}=z_k^\phi \in [0, 1]$ based on the current belief map $\tilde{R}$, where $J_{\text{satisfaction}} = 1$ when the task specification is fully satisfied and 0 if not satisfied at all. \

\subsubsection{Exploration}
To balance task execution with environment exploration, we incorporate an additional exploration objective into our optimization model. This objective encourages robots to visit areas of high uncertainty, thereby improving our belief about resource distribution over time.
The exploration objective is formulated as follows:
\begin{equation}
\label{eq: obj_exp}
J_{\text{exploration}} =
\sum_{q \in \mathcal{Q}}{ y(q) \cdot \Omega(q)},
\end{equation}
where $y(q)$ is a binary variable indicating whether location $q$ is visited. As defined earlier, the term $\Omega(q)$ represents the uncertainty at location $q$.
To ensure that $y(q)$ accurately represents location visits, we introduce the following constraints:
\begin{equation}
\label{eq: contr1}
y(q) \cdot n_g \geq \sum_{k=0}^{K-1} x_{q,g,k}, \quad \forall q \in \mathcal{Q}, g \in \mathcal{G},
\end{equation}
\begin{equation}
\label{eq: contr2}
\sum_{k=0}^{T-1} \sum_{g \in \mathcal{G}} x_{q,g,k} \geq y(q), \quad \forall q \in \mathcal{Q}.
\end{equation}
Here, $x_{q,g,k} \in \mathbb{B}$ indicates whether an agent with capability $g$ is at location $q$ at time step $k$. $K$ represents the time horizon in the $\texttt{ComputePlan}$, $\mathcal{G}$ is the set of all agent capabilities, and $n_g$ is the number of agents with capability $g$. 
The first constraint \eqref{eq: contr1}, ensures that $y(q)$ is set to 1 if any agent of any capability visits location $q$ at any time step. The second constraint \eqref{eq: contr2} guarantees that $y(q)$ is 0 if no agent visits location $q$ throughout the entire time horizon.
By incorporating this exploration objective and its associated constraints, our optimization model seeks to balance completing assigned tasks with exploring uncertain areas. This approach allows the team to progressively refine its understanding of the resource distribution while working towards mission objectives.

\subsubsection{MILP problem}
Now we can formulate the complete MILP problem that balances exploration of locations to improve information about resources distribution where the uncertainty is high and satisfy the mission with the current available information defined as follows
\begin{equation}
\label{eq: obj_partial}
\begin{aligned}
    \max_{}\quad &J_{\text{satisfaction}} + \alpha J_{exploration}\\
    & \text{robots and resources dynamics~\cite{cardona2024planning},} \\
    & \text{partial satisfaction encoding of } \phi \text{~\cite{cardona2022partial}. }
\end{aligned}
\end{equation}
where $\alpha$ is the exploration weight constant that balances exploration and task satisfaction.

\subsection{Belief Update}
We assume that an observation model is provided such that
a Kalman Filter \cite{thrun2002probabilistic} can be applied to estimate the environment's resources $\tilde{\mathcal{R}}_t \sim \mathcal{N}(\mu_t, \Sigma_t)$, where $\mu_t$ is the mean estimate and $\Sigma_t$ is the covariance matrix. 

Based on the $plan$ solved from the MILP in \eqref{eq: obj_partial}, the agents execute the $plan$ and collect a sequence of observations based on the camera $c_0$ and update $\Sigma_t$ using the Kalman filter. 
We generate a normalized variance map to capture the uncertainty for each location and guide future exploration.
The uncertainty for each location $q$ is calculated using the variances from the updated covariance matrix $\Sigma_t$:
\begin{equation}
\Omega(q) = \frac{\sum_{h \in \Resources} \alpha_{h}\sigma^2_r(q)}{\sum_{q' \in \mathcal{Q}} \sum_{h \in \Resources} \alpha_{h} \sigma^2_h(q')}
\end{equation}
where $\mathcal{Q}$ is the set of all locations, and $\sigma^2_h(q)$ is the variance of resource $h$ at location $q$, which corresponds to a diagonal element of $\Sigma_t$, $\alpha_h \in \mathbb{R}$ is the weight for different resources estimation variance. 
This formulation ensures that $\Omega(q)$ represents the relative uncertainty at location $q$, with higher values indicating greater uncertainty. The normalization ensures that $\sum_{q \in \mathcal{Q}} \Omega(q) = 1$.
The updated uncertainty map $\Omega$ is then used in the subsequent planning phase in the next iteration, specifically in the exploration objective in planning~\eqref{eq: obj_exp}.




\section{Computational Results}
\label{sec:results}
This section demonstrates the performance of the iterative approach for resource uncertainty. All computations of the following case studies were performed on a PC with 20 cores at 3.7 GHz with 64 GB of RAM.
We used Gurobi~\cite{gurobi} as the MILP solver. 
For encoding of CaTL specifications we used \href{https://github.com/erl-lehigh/PyTeLo}{PyTeLo}~\cite{cardona2023flexible} and ANTLRv4~\cite{parr2007definitive}, LOMAP~\cite{lomap} and networkx~\cite{SciPyProceedings_11} for transition system models of environments.

\noindent
\subsubsection*{Simulation specification} 
\begin{figure}
    \centering
\includegraphics[width=1\linewidth]{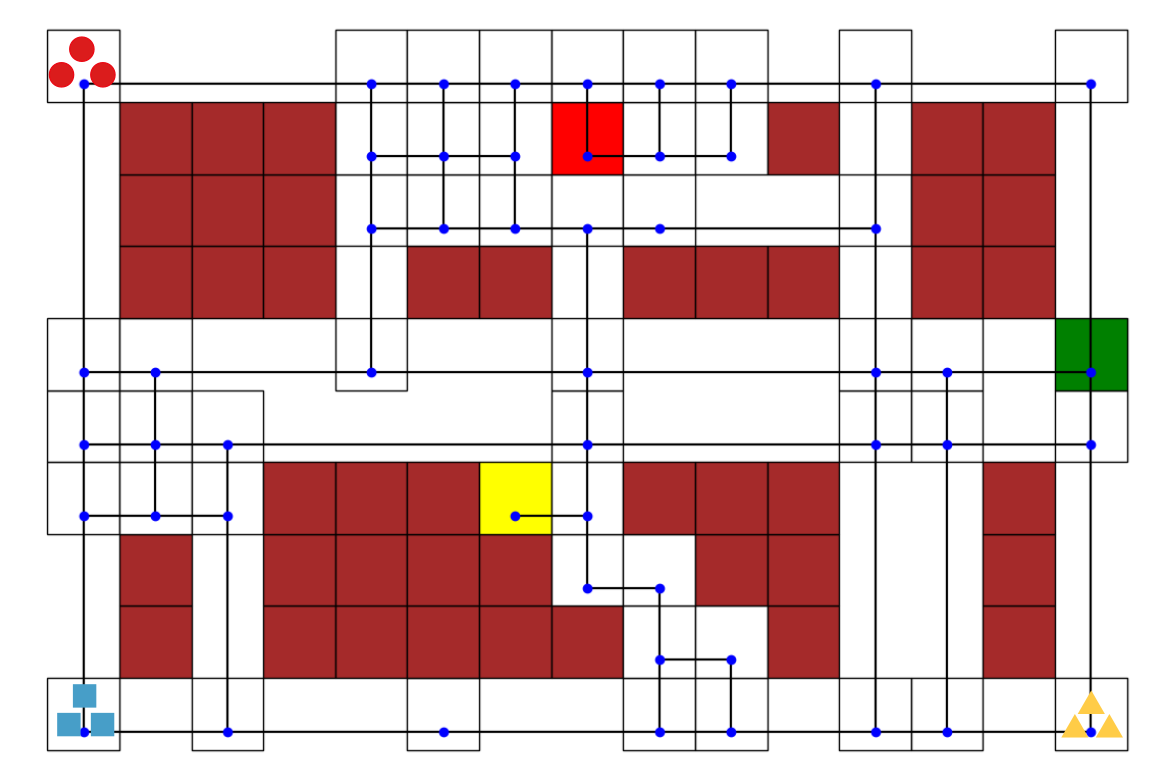}
\caption{Simulation Environment}
    \label{fig: env}
\end{figure}

The simulation environment is shown in Fig.~\ref{fig: env} which is a abstraction of transition system from Fig.~\ref{fig:problem_example}. It comprises a grid-based layout where brown areas represent impermissible regions, and interconnected lines denote travel edges between nodes. Each node is represented by a blue dot. The edge weights are proportional to their length, with the shortest edge weighting 1. 
We deploy three types of agents with corresponding initial locations and quantities as $a: (\texttt{bottom left}, 3), b: (\texttt{top left}, 3)$ and $c:(\texttt{bottom right}, 3)$. Two types of resources, $h_1$ and $ h_2$, are distributed in certain nodes, each resource with a total amount of $50$ in the environment, ensuring that maximum satisfaction to 100 percent is feasible. 
The specification is given as:
\begin{equation*}
    \begin{aligned}
        \phi &= \levent_{[0:30]}\phi_1 \land \levent_{[0:30]}\phi_2 \land  \levent_{[0:30]}\phi_3\\
        \phi_1 &= (2, \texttt{red}, {(a, 1), (b, 1)}, {(h_1, 15), (h_2, 15)})\\
        \phi_2 &= (1, \texttt{yellow}, {(c, 2), (a, 1), (b, 1)}, {(h_1, 15), (h_2, 15)})\\
        \phi_3 &= (1, \texttt{green}, {(c, 1), (a, 1), (b, 1)}, {(h_1, 10), (h_2, 10)})
    \end{aligned}
\end{equation*}

\begin{figure}[hbt!]

    \centering
    \subfloat[\label{subfig: satis}]{%
\includegraphics[width=0.8\linewidth]{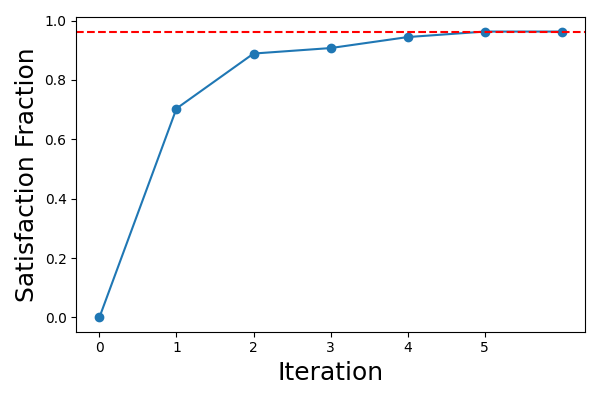}}
    \hspace{8pt}
    \centering
    \subfloat[\label{subfig: avg_error}]{\includegraphics[width=0.755\linewidth]{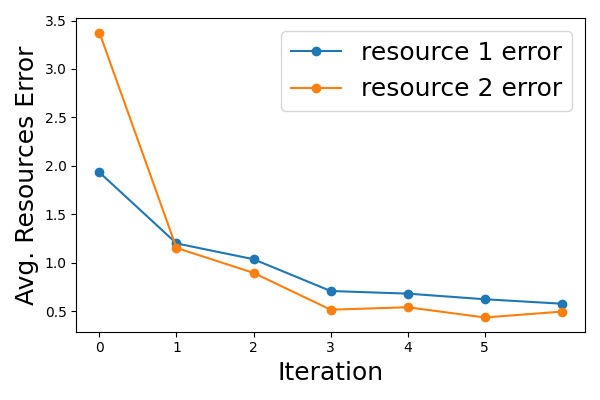}
       }
     \caption{Iterative performance (a) satisfaction fraction, (b) average resources estimation error.}
    \label{fig: main_exp}
\end{figure}

We start with exploration factor $\alpha = 1$ and a constant decaying factor of $0.8$ at each iteration as confidence grows.
We stop the iterative process when the satisfaction fraction converges. The result is shown in Fig.~\ref{fig: main_exp}. The satisfaction rate starts at $0$ since the initial belief map is initialized with $0$. As the estimation error of the resources decreases, shown in Fig.~\ref{subfig: avg_error}, the satisfaction fraction steadily converges to 100 percent in 7 iterations. The correlation between resource estimation error and satisfaction fraction underscores the effectiveness of environment exploration and task execution.

\subsubsection*{Run time analysis}

\begin{figure}[hbt!]
    \centering
    \includegraphics[width=1\linewidth]{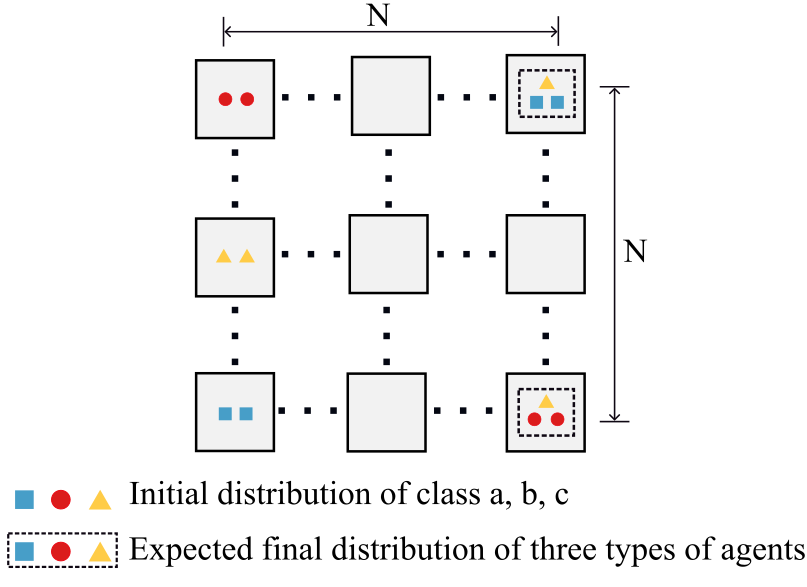}
    \caption{Run time experiment setup: Varying $M$ (showing $M = 2$) number of three types of agents employed in varying size of the $N$ by $N$ grid environment}
    \label{fig: scal_fig}
\end{figure}

The experimental setup evaluates the performance of the iterative approach across a range of grid sizes and agent populations. The setup is shown in Fig.~\ref{fig: scal_fig}. The environment is a $N \times N$ grid where $N \in \{5, 6, 7, 8\}$. The transition weight for connected nodes is 1. Three classes of robot is used in this environment to accumulate two types of resources.
The three robot classes have standard deviations of 0.5, 0.8, 0.5 respectively when gathering resources.
Each class contains $M$ agents, where $M \in \{1, 2, 3, 4, 5\}$. 

Initial distribution of total $3\times M$ agents are:
\begin{itemize}
\item Type $a$ ($M$ agents): Bottom-left corner
\item Type $b$ ($M$ agents): Top-left corner
\item Type $c$ ($M$ agents): Middle of left column
\end{itemize}

The task specification is given as:
\begin{equation*}
\begin{aligned}
\phi &= \Diamond_{I} \phi_1 \land \Diamond_{I} \phi_2, \qquad I = [0, 3\times 
 N], \\
\phi_1 &= (1, \text{Bottom Right}, {(a, M), (c, \ceil{M/2})}, {(h_1, 10 \times M)}),\\
\phi_2 &= (1, \text{Top Right}, {(b, M), (c, \floor{M, 2})}, {(h_2, 10 \times M)})
\end{aligned}
\end{equation*}
The total amount of each resource is $15\times M$ and is randomly distributed in the environment. The overall specification requires all type $a$ and half of type $c$ agents move to top right corner and all type $b$ and half of type $c$ agents to the right bottom corner with a specific amount of resources demand. 
This setup ensures that the specification and resource distribution scale with the size of the environment and the number of agents.

\begin{figure}[hbt!]
    \centering
    \subfloat[\label{subfig: scal}]{%
\includegraphics[width=0.85\linewidth]{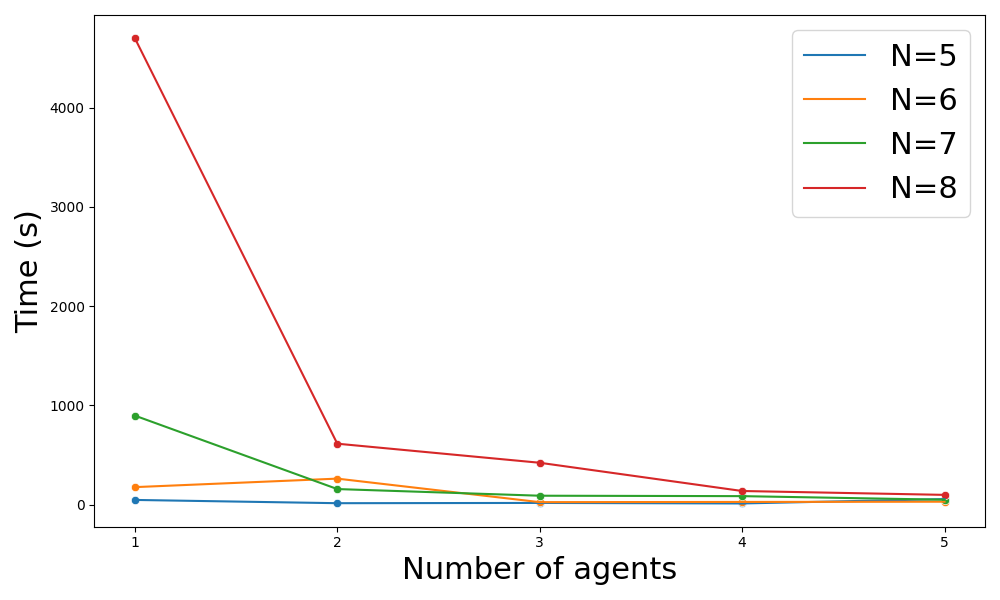}}
    \hspace{8pt}
    \centering
    \subfloat[\label{subfig: avg_itr}]{\includegraphics[width=0.85\linewidth]{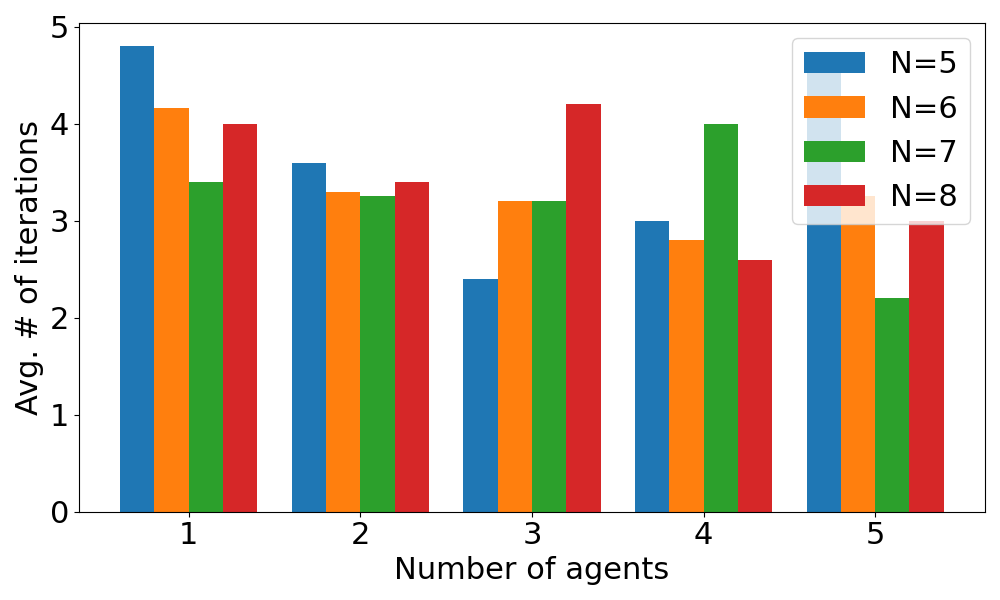}
       }
     \caption{Scalability analysis (a) run time performance, (b)average number of iterations for convergence}
    \label{fig: scal}
\end{figure}

Fig.~\ref{subfig: scal} demonstrates the run time results of this analysis, the iteration process stops when the objective of the satisfaction fraction converges. Each data point is the average of 5 runs. The result shows that the runtime decreases consistently with the increasing number of agents for different sizes of environments. This suggests that more agents in the environment facilitate faster solution discovery by the optimization process. Furthermore, the result also shows that the run time discrepancy between large and smaller environments decreases as the number of agents increases.

Fig.~\ref{subfig: avg_itr} shows the average number of iterations required for convergence across various environmental configurations and agent populations. All scenarios achieve a 100 percent satisfaction rate. The consistently low number of iterations demonstrates the algorithm's efficiency and adaptability across diverse problem scales.

\section{Conclusions and Future Work}
\label{sec:conclusion}
This work introduces a framework for simultaneous exploration and task fulfillment by a team of heterogeneous robots operating under Capability Temporal Logic (CaTL) specifications. 
We propose an iterative approach that enables the robots to progressively refine their understanding of the environment while pursuing mission objectives. 
The algorithm's iterative nature ensures continuous updates to the belief map, allowing decisions to be made based on the most up-to-date information and thereby maximizing task satisfaction. 
We demonstrate that our method effectively balances exploration and task execution within a limited number of iterations. 
In future work, we aim to extend the framework to account for uncertainty and failure in transition edges and to validate its functionality and performance on physical robotic systems.
\bibliographystyle{ieeetr}
\bibliography{references.bib}

\begin{thebibliography}{10}

\bibitem{cardona2019robot}
G.~A. Cardona and J.~M. Calderon, ``Robot swarm navigation and victim detection using rendezvous consensus in search and rescue operations,'' {\em Applied Sciences}, vol.~9, no.~8, p.~1702, 2019.

\bibitem{gregory2016application}
J.~Gregory, J.~Fink, E.~Stump, J.~Twigg, J.~Rogers, D.~Baran, N.~Fung, and S.~Young, ``Application of multi-robot systems to disaster-relief scenarios with limited communication,'' in {\em Field and Service Robotics: Results of the 10th International Conference}, pp.~639--653, Springer, 2016.

\bibitem{cardona2022planning}
G.~A. Cardona, D.~Salda{\~n}a, and C.-I. Vasile, ``Planning for modular aerial robotic tools with temporal logic constraints,'' in {\em IEEE Conference on Decision and Control (CDC)}, pp.~2878--2883, IEEE, 2022.

\bibitem{weisbin2000nasa}
C.~R. Weisbin and G.~Rodriguez, ``Nasa robotics research for planetary surface exploration,'' {\em IEEE Robotics \& Automation Magazine}, vol.~7, no.~4, pp.~25--34, 2000.

\bibitem{di2010autonomous}
D.~Di~Paola, A.~Milella, G.~Cicirelli, and A.~Distante, ``An autonomous mobile robotic system for surveillance of indoor environments,'' {\em International Journal of Advanced Robotic Systems}, vol.~7, no.~1, p.~8, 2010.

\bibitem{parascho2023construction}
S.~Parascho, ``Construction robotics: From automation to collaboration,'' {\em Annual Review of Control, Robotics, and Autonomous Systems}, vol.~6, no.~1, pp.~183--204, 2023.

\bibitem{zaheer2016aerial}
Z.~Zaheer, A.~Usmani, E.~Khan, and M.~A. Qadeer, ``Aerial surveillance system using uav,'' in {\em 2016 thirteenth international conference on wireless and optical communications networks (WOCN)}, pp.~1--7, IEEE, 2016.

\bibitem{fang2024continuous}
A.~Fang, T.~Yin, J.~Lin, and H.~Kress-Gazit, ``Continuous execution of high-level collaborative tasks for heterogeneous robot teams,'' {\em arXiv preprint arXiv:2406.18019}, 2024.

\bibitem{jing2018accomplishing}
G.~Jing, T.~Tosun, M.~Yim, and H.~Kress-Gazit, ``Accomplishing high-level tasks with modular robots,'' {\em Autonomous Robots}, vol.~42, pp.~1337--1354, 2018.

\bibitem{jing2016end}
G.~Jing, T.~Tosun, M.~Yim, and H.~Kress-Gazit, ``An end-to-end system for accomplishing tasks with modular robots.,'' in {\em Robotics: Science and systems}, vol.~2, 2016.

\bibitem{buyukkocak2023energy}
A.~T. Buyukkocak, D.~Aksaray, and Y.~Yaz{\i}c{\i}o{\u{g}}lu, ``Energy-aware planning of heterogeneous multi-agent systems for serving cooperative tasks with temporal logic specifications,'' in {\em 2023 IEEE/RSJ International Conference on Intelligent Robots and Systems (IROS)}, pp.~8659--8665, IEEE, 2023.

\bibitem{tumova2016multi}
J.~Tumova and D.~V. Dimarogonas, ``Multi-agent planning under local ltl specifications and event-based synchronization,'' {\em Automatica}, vol.~70, pp.~239--248, 2016.

\bibitem{kantaros2020stylus}
Y.~Kantaros and M.~M. Zavlanos, ``Stylus*: A temporal logic optimal control synthesis algorithm for large-scale multi-robot systems,'' {\em The International Journal of Robotics Research}, vol.~39, no.~7, pp.~812--836, 2020.

\bibitem{kress2009temporal}
H.~Kress-Gazit, G.~E. Fainekos, and G.~J. Pappas, ``Temporal-logic-based reactive mission and motion planning,'' {\em IEEE transactions on robotics}, vol.~25, no.~6, pp.~1370--1381, 2009.

\bibitem{sun2022multi}
D.~Sun, J.~Chen, S.~Mitra, and C.~Fan, ``Multi-agent motion planning from signal temporal logic specifications,'' {\em IEEE Robotics and Automation Letters}, vol.~7, no.~2, pp.~3451--3458, 2022.

\bibitem{gundana2021event}
D.~Gundana and H.~Kress-Gazit, ``Event-based signal temporal logic synthesis for single and multi-robot tasks,'' {\em IEEE Robotics and Automation Letters}, vol.~6, no.~2, pp.~3687--3694, 2021.

\bibitem{lindemann2019coupled}
L.~Lindemann, J.~Nowak, L.~Sch{\"o}nb{\"a}chler, M.~Guo, J.~Tumova, and D.~V. Dimarogonas, ``Coupled multi-robot systems under linear temporal logic and signal temporal logic tasks,'' {\em IEEE Transactions on Control Systems Technology}, vol.~29, no.~2, pp.~858--865, 2019.

\bibitem{sewlia2023maps}
M.~Sewlia, C.~K. Verginis, and D.~V. Dimarogonas, ``Maps2: Multi-robot anytime motion planning under signal temporal logic specifications,'' {\em arXiv preprint arXiv:2309.05632}, 2023.

\bibitem{buyukkocak2021planning}
A.~T. Buyukkocak, D.~Aksaray, and Y.~Yaz{\i}c{\i}o{\u{g}}lu, ``Planning of heterogeneous multi-agent systems under signal temporal logic specifications with integral predicates,'' {\em IEEE Robotics and Automation Letters}, vol.~6, no.~2, pp.~1375--1382, 2021.

\bibitem{Jones2019ScRATCHS}
A.~Jones, K.~Leahy, C.~Vasile, S.~Sadraddini, Z.~Serlin, R.~Tron, and C.~Belta, ``Scratchs: Scalable and robust algorithms for task-based coordination from high-level specifications,'' in {\em International Symposium of Robotics Research}, pp.~224--241, 2019.

\bibitem{leahy2021scalable}
K.~Leahy, Z.~Serlin, C.-I. Vasile, A.~Schoer, A.~M. Jones, R.~Tron, and C.~Belta, ``Scalable and robust algorithms for task-based coordination from high-level specifications (scratches),'' {\em IEEE Transactions on Robotics}, vol.~38, no.~4, pp.~2516--2535, 2021.

\bibitem{leahy2022fast}
K.~Leahy, A.~Jones, and C.-I. Vasile, ``Fast decomposition of temporal logic specifications for heterogeneous teams,'' {\em IEEE Robotics and Automation Letters}, vol.~7, no.~2, pp.~2297--2304, 2022.

\bibitem{cardona2024planning}
G.~A. Cardona and C.-I. Vasile, ``Planning for heterogeneous teams of robots with temporal logic, capability, and resource constraints,'' {\em The International Journal of Robotics Research}, p.~02783649241247285.

\bibitem{sadigh2016safe}
D.~Sadigh and A.~Kapoor, ``Safe control under uncertainty with probabilistic signal temporal logic,'' in {\em Proceedings of Robotics: Science and Systems XII}, 2016.

\bibitem{schillinger2018auctioning}
P.~Schillinger, M.~B{\"u}rger, and D.~V. Dimarogonas, ``Auctioning over probabilistic options for temporal logic-based multi-robot cooperation under uncertainty,'' in {\em 2018 IEEE International Conference on Robotics and Automation (ICRA)}, pp.~7330--7337, IEEE, 2018.

\bibitem{guo2023hierarchical}
M.~Guo, T.~Liao, J.~Wang, and Z.~Li, ``Hierarchical motion planning under probabilistic temporal tasks and safe-return constraints,'' {\em IEEE Transactions on Automatic Control}, vol.~68, no.~11, pp.~6727--6742, 2023.

\bibitem{jha2018safe}
S.~Jha, V.~Raman, D.~Sadigh, and S.~A. Seshia, ``Safe autonomy under perception uncertainty using chance-constrained temporal logic,'' {\em Journal of Automated Reasoning}, vol.~60, pp.~43--62, 2018.

\bibitem{cai2021probabilistic}
M.~Cai, K.~Leahy, Z.~Serlin, and C.-I. Vasile, ``Probabilistic coordination of heterogeneous teams from capability temporal logic specifications,'' {\em IEEE Robotics and Automation Letters}, vol.~7, no.~2, pp.~1190--1197, 2021.

\bibitem{liang2024iterative}
K.~Liang, G.~A. Cardona, and C.-I. Vasile, ``An iterative approach for heterogeneous multi-agent route planning with temporal logic goals and travel duration uncertainty,'' in {\em 2024 IEEE International Conference on Robotics and Automation (ICRA)}, pp.~257--263, IEEE, 2024.

\bibitem{maler2004}
O.~Maler and D.~Nickovic, ``Monitoring temporal properties of continuous signals,'' in {\em Formal Techniques, Modelling and Analysis of Timed and Fault-Tolerant Systems}, pp.~152--166, Springer, 2004.

\bibitem{liu2023robust}
W.~Liu, K.~Leahy, Z.~Serlin, and C.~Belta, ``Robust multi-agent coordination from catl+ specifications,'' in {\em 2023 American Control Conference (ACC)}, pp.~3529--3534, IEEE, 2023.

\bibitem{cardona2022partial}
G.~A. Cardona and C.-I. Vasile, ``{Partial Satisfaction of Signal Temporal Logic Specifications for Coordination of Multi-robot Systems},'' in {\em Workshop on the Algorithmic Foundations of Robotics}, pp.~223--238, Springer, 2022.

\bibitem{cardona2023preferences}
G.~A. Cardona and C.-I. Vasile, ``Preferences on partial satisfaction using weighted signal temporal logic specifications,'' in {\em 2023 European Control Conference (ECC)}, pp.~1--6, IEEE, 2023.

\bibitem{thrun2002probabilistic}
S.~Thrun, ``Probabilistic robotics,'' {\em Communications of the ACM}, vol.~45, no.~3, pp.~52--57, 2002.

\bibitem{gurobi}
L.~Gurobi~Optimization, ``Gurobi optimizer reference manual,'' 2020.

\bibitem{cardona2023flexible}
G.~A. Cardona, K.~Leahy, M.~Mann, and C.-I. Vasile, ``A flexible and efficient temporal logic tool for python: Pytelo,'' {\em arXiv preprint arXiv:2310.08714}, 2023.

\bibitem{parr2007definitive}
T.~Parr, {\em The definitive ANTLR reference: building domain-specific languages}.
\newblock Pragmatic Bookshelf, 2007.

\bibitem{lomap}
C.-I. Vasile and A.~Ulusoy, ``Ltl optimal multi-agent planner (lomap).'' \url{https://github.com/wasserfeder/lomap}, 2024.

\bibitem{SciPyProceedings_11}
A.~A. Hagberg, D.~A. Schult, and P.~J. Swart, ``Exploring network structure, dynamics, and function using networkx,'' in {\em Proceedings of the 7th Python in Science Conference} (G.~Varoquaux, T.~Vaught, and J.~Millman, eds.), (Pasadena, CA USA), pp.~11 -- 15, 2008.

\end{thebibliography}
\end{document}